\documentclass[10pt,twocolumn,letterpaper]{article}

\usepackage[pagenumbers]{cvpr}      

\usepackage{abbrv}
\usepackage[dvipsnames]{xcolor}
\usepackage{url}
\usepackage{tikz}
\usepackage{comment}
\usepackage{amsthm}
\usepackage{amsmath,amssymb,bbm}
\usepackage{colortbl}
\usepackage{epsfig}
\usepackage{caption}
\usepackage{soul}
\usepackage[utf8]{inputenc}
\usepackage{listings}
\usepackage[T1]{fontenc}
\usepackage{booktabs}
\usepackage{amsfonts}
\usepackage{nicefrac}
\usepackage{microtype}
\usepackage{threeparttable}
\usepackage{algorithm-mine}
\usepackage{numprint}
\usepackage{siunitx}
\usepackage{etoolbox}
\usepackage{cancel}
\newcommand{\ubold}{\fontseries{b}\selectfont}
\robustify\ubold
\usepackage{wrapfig}
\usepackage{algpseudocode}
\usepackage[skip=2pt,font=footnotesize,labelfont=footnotesize]{subcaption}
\usepackage{pifont}
\usepackage{lineno}
\usepackage{array,multirow,graphicx}
\usepackage{tabularx}
\usepackage{listings}
\usepackage[para]{footmisc}

\usepackage[accsupp]{axessibility}
\usepackage{pgfplots}
\lstdefinelanguage{json}{
    basicstyle=\normalfont\ttfamily,
    numbers=left,
    numberstyle=\scriptsize,
    stepnumber=1,
    numbersep=8pt,
    showstringspaces=false,
    breaklines=true,
    frame=lines,
    backgroundcolor=\color{background},
    literate=
     *{0}{{{\color{numb}0}}}{1}
      {1}{{{\color{numb}1}}}{1}
      {2}{{{\color{numb}2}}}{1}
      {3}{{{\color{numb}3}}}{1}
      {4}{{{\color{numb}4}}}{1}
      {5}{{{\color{numb}5}}}{1}
      {6}{{{\color{numb}6}}}{1}
      {7}{{{\color{numb}7}}}{1}
      {8}{{{\color{numb}8}}}{1}
      {9}{{{\color{numb}9}}}{1}
      {:}{{{\color{punct}{:}}}}{1}
      {,}{{{\color{punct}{,}}}}{1}
      {\{}{{{\color{delim}{\{}}}}{1}
      {\}}{{{\color{delim}{\}}}}}{1}
      {[}{{{\color{delim}{[}}}}{1}
      {]}{{{\color{delim}{]}}}}{1},
}

\pgfplotsset{compat=1.18}

\definecolor{cvprblue}{rgb}{0.21,0.49,0.74}
\usepackage[pagebackref,breaklinks,colorlinks,allcolors=cvprblue]{hyperref}

\definecolor{keyword}{rgb}{.224,.451,.686}
\definecolor{tabhighlight}{HTML}{e5e5e5}
\definecolor{lightblue}{rgb}{0.63, 0.79, 0.95}
\definecolor{babypink}{rgb}{0.96, 0.76, 0.76}
\definecolor{skyblue}{rgb}{0.53, 0.81, 0.92}
\definecolor{wheat}{rgb}{0.96, 0.87, 0.7}
\definecolor{delim}{RGB}{20,105,176}
\colorlet{punct}{red!60!black}
\colorlet{numb}{magenta!60!black}

\newcommand{\keyword}[1]{\textcolor{keyword}{#1}}

\newcommand{\keywordtwo}[1]{\textcolor{Green}{#1}}
\newcommand{\keywordtri}[1]{\textcolor{Purple}{#1}}
\newcommand{\cmark}{\textcolor{OliveGreen}{\ding{51}}}
\newcommand{\xmark}{\textcolor{BrickRed}{\ding{55}}}

\def\paperID{8977} 
\def\confName{CVPR}
\def\confYear{2025}

\title{HyperGLM: HyperGraph for Video Scene Graph Generation and Anticipation}

\author{Trong-Thuan Nguyen$^{1\dag}$, Pha Nguyen$^{1\dag}$, Jackson Cothren$^{1}$, Alper Yilmaz$^{2}$, Khoa Luu$^{1}$ \\
\small $^{1}$University of Arkansas \quad
\small $^{2}$Ohio State University\\
\tt\small $^{1}$\{thuann, panguyen, jcothre, khoaluu\}@uark.edu \quad $^{2}$yilmaz.15@osu.edu \\
\small \tt \href{https://uark-cviu.github.io/projects/HyperGLM/}{uark-cviu.github.io/projects/HyperGLM}
\\
\small \tt \href{}{}
}

\begin{document}

\twocolumn[{
 \renewcommand\twocolumn[1][]{#1}%
 \maketitle
 \begin{center}
 \centering
 \captionsetup{type=figure}
 \includegraphics[width=.9\textwidth]{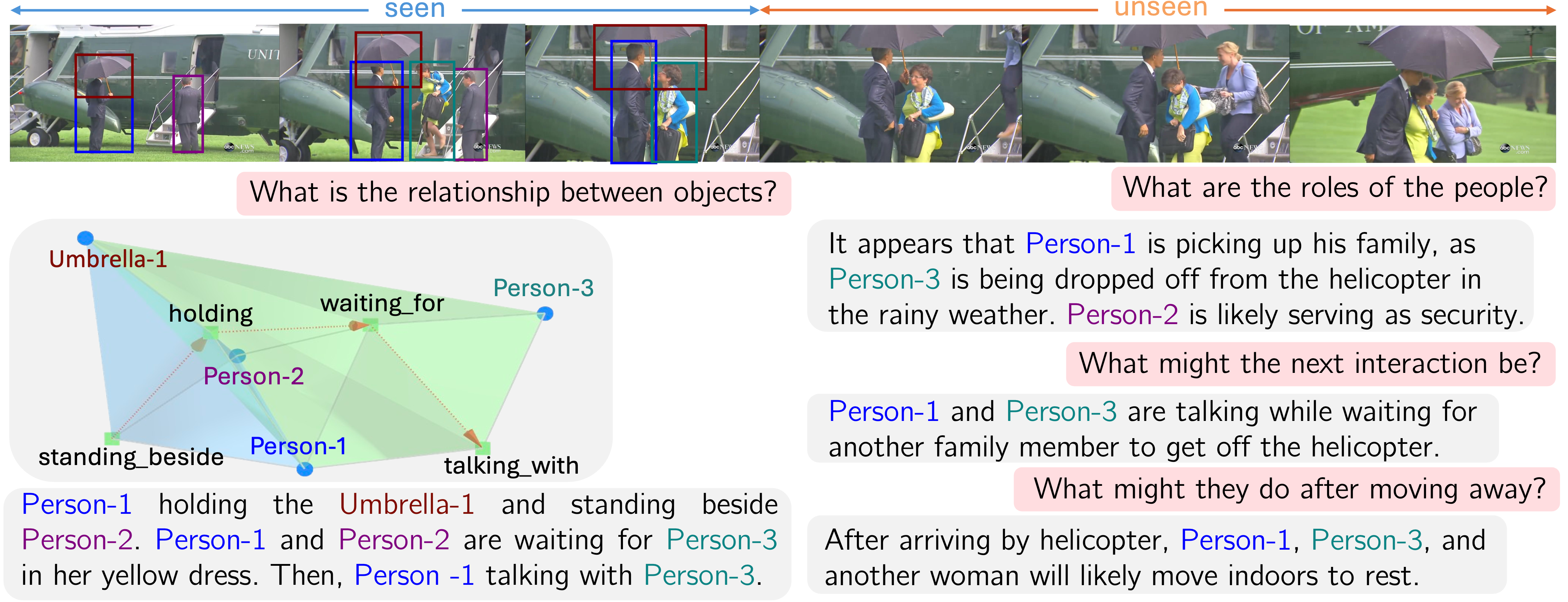}
 \captionof{figure}{Our HyperGLM framework supports Video Scene Graph Generation, Anticipation, and Reasoning. HyperGLM constructs scene graphs from observed video frames and predicts relationships in unseen frames by leveraging a unified hypergraph for temporal modeling and comprehensive understanding.}
 \label{fig:motivation}
 \end{center}
 }]

 \if TT\insert\footins{\footnotesize{$\dag$ Equal contribution.}}\fi

\begin{abstract}
Multimodal LLMs have advanced vision-language tasks but still struggle with understanding video scenes. To bridge this gap, Video Scene Graph Generation (VidSGG) has emerged to capture multi-object relationships across video frames. However, prior methods rely on pairwise connections, limiting their ability to handle complex multi-object interactions and reasoning. To this end, we propose the Multimodal Large Language Models (LLMs) on a Scene HyperGraph (HyperGLM), promoting reasoning about multi-way interactions and higher-order relationships. Our approach uniquely integrates entity scene graphs, which capture spatial relationships between objects, with a procedural graph that models their causal transitions, forming a unified HyperGraph. Significantly, HyperGLM enables reasoning by injecting this unified HyperGraph into LLMs. Additionally, we introduce a new Video Scene Graph Reasoning (VSGR) dataset featuring 1.9M frames from third-person, egocentric, and drone views and support five tasks. Empirically, HyperGLM consistently outperforms state-of-the-art methods, effectively modeling and reasoning complex relationships in diverse scenes.
\end{abstract}


\section{Introduction}\label{sec:intro}\vspace{-0.25\baselineskip}










In recent years, Multimodal Large Language Models (LLMs)~\cite{fei2024video, chen2024videollm} have set new benchmarks, with vision-language models excelling in diverse multimodal tasks. However, fully understanding dynamic video scenes remains a significant challenge for applications like autonomous driving, intelligent surveillance, human-object interaction, and multimedia analysis. Towards this goal, Video Scene Graph Generation (VidSGG)~\cite{ji2020action, yang2023panoptic} has emerged as a critical task for capturing multi-object relationships across video frames. In particular, VidSGG enables high-level tasks such as event forecasting~\cite{thakur2024graph, wang2023memory, peddi2024towards}, video captioning~\cite{seo2022end, shen2023accurate, ko2023meltr}, and video question answering~\cite{maaz2023video, lin2023video, song2024moviechat, jin2024chat} by constructing detailed representations of entities and their interactions.

However, prior VidSGG methods and datasets have limited capacity for comprehensive video understanding. Traditional scene graph-based methods~\cite{cong2021spatial, teng2021target, nguyen2024hig, nguyen2024cyclo} only model pairwise object relationships within single frames, making it challenging to capture higher-order relationships and temporal dependencies in real-world scenarios. Additionally, existing benchmark datasets~\cite{ji2020action, yang2023panoptic, nguyen2024hig, nguyen2024cyclo} focus is primarily confined to Scene Graph Generation (SGG) and Scene Graph Anticipation (SGA) tasks, lacking annotations for reasoning tasks such as Video Question Answering (VQA), Video Captioning (VC), and Relation Reasoning (RR).

In this paper, we propose the Multimodal LLMs on a Scene HyperGraph (HyperGLM) approach to promote \textit{reasoning about multi-way interactions and higher-order relationships} through a unified HyperGraph and LLMs, as illustrated in Fig.~\ref{fig:motivation}. To achieve this goal, we uniquely incorporate \textit{entity scene graphs}, which capture spatial relationships between objects, with \textit{a procedural graph} that models their causal transitions across video frames, forming \textit{a unified HyperGraph}, as shown in Fig.~\ref{fig:compare}. Significantly, our HyperGLM approach allows reasoning by injecting this unified HyperGraph into LLMs. In addition, we introduce a novel Video Scene Graph Reasoning (VSGR) dataset, comprising 1.9 million video frames surpassing existing benchmark datasets~\cite{ji2020action, yang2023panoptic, wu2024sportshhi, nguyen2024hig, nguyen2024cyclo} in scale and annotation depth. Specifically, our VSGR dataset includes videos from third-person, egocentric, and drone perspectives. It supports five tasks: Scene Graph Generation, Scene Graph Anticipation, Video Question Answering, Video Captioning, and Relation Reasoning. Notably, our VSGR dataset introduces a new \textit{Relation Reasoning} task, setting it apart from existing video scene graph datasets, as shown in the comparison in Table~\ref{tab:dataset_comparison}.

\noindent \textbf{Contributions of this Work.} This work presents three contributions to advance Video Scene Graph Generation. First, we introduce Multimodal LLMs on a Scene HyperGraph, leveraging hyperedges and LLMs for \textit{reasoning about multi-way interactions} and \textit{higher-order relationships}. Second, we develop a new Video Scene Graph Reasoning dataset, surpassing existing scale and annotation depth benchmarks. Our VSGR dataset primarily supports five tasks within diverse video scenes. Finally, the proposed HyperGLM consistently outperforms state-of-the-art methods across all five tasks.

\section{Related Work}\label{sec:related_work}
In this section, we review advances in scene graph generation and hypergraph applications in computer vision, then discuss existing limitations and the advantages of our approach.

\subsection{Scene Graph Generation}\label{sec:related_sgg}
Scene Graph Generation has significantly advanced with transformer-based models~\cite{yang2022panoptic, li2022sgtr, kundu2023ggt, cong2023reltr, im2024egtr, hayder2024dsgg} that have become benchmarks due to their efficiency and state-of-the-art performance. Recent work~\cite{chen2023more, jin2023fast, nag2023unbiased} focuses on reducing bias and enhancing mean recall for rare predicates by integrating external knowledge and applying unbiased contextual augmentation, particularly in dynamic video contexts. Open-vocabulary methods~\cite{he2022towards, li2024pixels} supported by vision-language models handle unseen object and relationship classes, improving generalization. Additionally, generative models (e.g., diffusion-based methods~\cite{farshad2023scenegenie, zhai2024commonscenes}) leverage scene graphs for efficient image and scene synthesis. Moreover, spatial-temporal methods~\cite{cong2021spatial, teng2021target, nguyen2024hig, nguyen2024cyclo, peddi2024towards} effectively capture dynamic object relationships in videos. Recently, Large Language Models~\cite{kim2024llm4sgg} have been utilized to enhance triplet extraction and alignment in weakly supervised SGG.

\subsection{HyperGraphs}
HyperGraphs have been adopted in computer vision to model complex multimodal data and capture higher-order relationships. Unlike traditional graph-based approaches, HyperGraphs connect multiple nodes through hyperedges, enabling multi-way relationships. They enhance Graph Neural Networks (GNNs)~\cite{zhang2022deep, gao2022hgnn+} by allowing the modeling of more sophisticated interactions. Recent advancements, such as HyperGraph Convolution~\cite{bai2021HyperGraph} and HyperGraph Attention~\cite{kim2020HyperGraph}, have further improved GNNs by capturing relationships beyond simple pairwise connections. Therefore, HyperGraph-based models effectively handle temporal dependencies and complex interactions, significantly boosting performance in tasks like accident anticipation~\cite{thakur2024graph}, group activity recognition~\cite{zhu2024dynamical, li2022multi}, and video question answering~\cite{wang2024multi, xu2022modeling, urooj2023learning}.

\subsection{Discussion}

\noindent\textbf{Limitations in Prior Methods.} The methods introduced in Section~\ref{sec:related_sgg}, based on \textit{Progressive Feature Fusion}~\cite{yang2022panoptic, shang2021video}, \textit{Batch-Progressive Transformer Encoding}~\cite{li2022sgtr, kundu2023ggt, cong2023reltr, im2024egtr, hayder2024dsgg}, \textit{Spatial-Temporal Context Integration}~\cite{cong2021spatial, teng2021target, nguyen2024hig}, and \textit{Memory-Guided Temporal Consistency}~\cite{deng2022hierarchical, nag2023unbiased, nguyen2024cyclo}, have advanced VidSGG. However, these methods struggle to model \textit{higher-order relationships} and complex temporal dynamics. Specifically, \textit{Progressive Feature Fusion} and \textit{Batch-Progressive Transformer Encoding} are limited in capturing long-term temporal dependencies, with the former lacking long-term context due to frame-by-frame processing and the latter only addressing short-term dependencies. Similarly, \textit{Spatial-Temporal Context Integration} and \textit{Memory-Guided Temporal Consistency} inadequately represent multi-object interactions across video frames and insufficiently capture the temporal evolution required for higher-order relationships.

\begin{figure}
\centering
\includegraphics[width=0.5\textwidth]{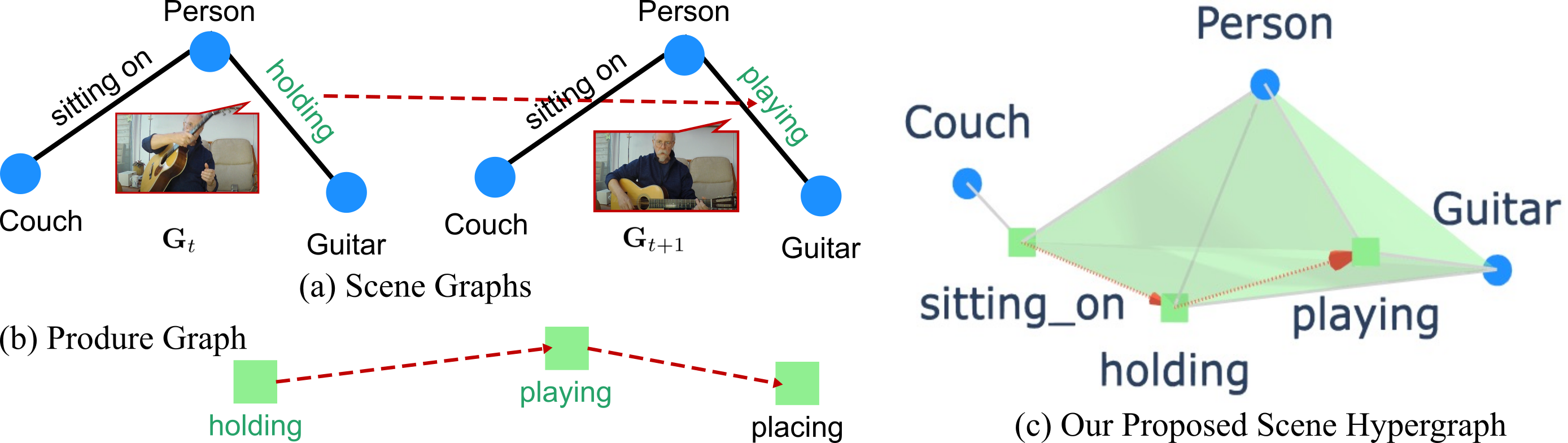}
\caption{(a) To model the temporal transition, a simple approach can be using two scene graphs $\mathbf{G_{t}}$ and $\mathbf{G_{t + 1}}$. (b) Another procedure graph can present this temporal modeling. (c) Our unified HyperGraph in Fig.~\ref{fig:compare}c integrates both \textit{entity scene graph} to capture spatial relationships and the \textit{procedural graph} to model the temporal evolution. \textit{HyperEdge} represents \textcolor{Blue}{\textit{person}} \textcolor{ForestGreen}{\textit{sitting on}} \textcolor{Blue}{\textit{couch}}, \textcolor{ForestGreen}{\textit{holding}}, \textcolor{red!75}{then} \textcolor{ForestGreen}{\textit{playing}} \textcolor{Blue}{\textit{guitar}}, whereas \textit{holding \textcolor{red!75}{$\rightarrow$} playing} describes a chain of interactions. HyperGraph is presented in 3D.} 
\label{fig:compare}
\vspace{-1\baselineskip}
\end{figure}

\noindent\textbf{Advantages of Our Approach.} Our HyperGLM approach, Multimodal LLMs on a Scene HyperGraph, promotes \textit{reasoning about multi-way interactions and high-order relationships}. As illustrated in Fig.~\ref{fig:compare}, HyperGLM enhances the model's ability to interpret complex relationships and anticipate intricate video dynamics. Towards this goal, we uniquely integrate \textit{entity scene graphs}, which is introduced in~\cref{sec:prob_form} to capture spatial interactions between objects in each frame and a \textit{procedural graph}, which is presented in~\cref{sec:vsh} to model their causal evolution. In our unified approach, hyperedges connect multiple nodes to capture higher-order relationships distinguished from traditional pairwise methods~\cite{cong2021spatial, teng2021target, nguyen2024hig, nguyen2024cyclo}. In addition, the procedural graph enables \textit{multi-step transitions} for anticipating future interactions or relationships, and \textit{reduces bias} by generalizing infrequent relationship categories.
Furthermore, our HyperGLM approach leverages fundamental mathematical properties: \textit{permutation equivariance} ensures that the HyperGraph structure remains consistent under any permutation of node labels, and \textit{invariance to hyperedge order} preserves semantic meanings regardless of the node visit order during random walks which is defined in Alg.~\ref{alg:random_walk}. The theoretical foundations and mathematical properties are detailed in the \textbf{Appendices}.

\section{Problem Formulation}\label{sec:prob_form}
In this section, we define two tasks, including Scene Graph Generation (SGG) and Scene Graph Anticipation (SGA). Graph vertex sequence is represented as $ \{V_{\mathbf{G}} \mid 1 \leq t \leq T\} $, where each set of vertex $ V_{\mathbf{G}} $ contains object features, bounding boxes, and categories. The scene graph for each frame $ t $, denoted by $ \mathbf{G}_t = \big(V_{\mathbf{G}}, E_{\mathbf{G}} = \{r^{ij} \mid 1 \leq i < j \leq |V_{\mathbf{G}}| \}\big) $, consists of all pairwise relationships between objects, with $ r^{ij} $ representing the relationship category between $ v_i $ and $ v_j $.

We aim to develop a process $ p_\theta: (V_{\mathbf{G}} \times V_{\mathbf{G}}) \to r^{ij} $ to predict the relationship $ r^{ij} $ between each object pair $ (v_t^i, v_t^j) $ in $ V_{\mathbf{G}} $. We define task-specific queries $ \mathbf{Q}_{\text{SGG}} $, and $ \mathbf{Q}_{\text{SGA}} $ to direct \textit{our unified model} to perform on each specific task. The objective for each task is to minimize the negative likelihood of the predicted scene graph $ \mathbf{G}_t $ to the truth predicate set $ Y_t = \{y^{ij} \mid 1 \leq i < j \leq |V_{\mathbf{G}}|\}$ on categories indexed by $k$.




\noindent\textbf{Scene Graph Generation (SGG)} generates the scene graph for each frame $ t $ from $ t = 1 $ to $ t = T $, which is defined as:
\begin{equation}\label{eq:sgg_formulation}
\small
\min_\theta \mathbb{E}_{\mathbf{G}_t, Y_t} \Big[ - \sum_{(v_t^{i, j})} \sum_{k} \big(y_k^{ij} \log p_\theta(r^{ij} \mid \mathbf{Q}_{\text{SGG}})_k\big) \Big]
\end{equation}
\noindent\textbf{Scene Graph Anticipation (SGA)} anticipates the scene graphs for future frames, generating predictions $ \mathbf{G}_{t+n} $, where $n$ denotes the anticipation horizon, formulated as:
\begin{equation}\label{eq:sga_formulation}
\small
\min_\theta \mathbb{E}_{\keyword{\mathbf{G}_{\leq t}}, Y_t} \Big[ - \sum_{(v_{t\keyword{+n}}^{i, j})} \sum_{k} \big(y_k^{ij} \log p_\theta(r^{ij} \mid \mathbf{Q}_{\text{SGA}})_k\big) \Big]
\end{equation}
In Eqns.~\eqref{eq:sgg_formulation} and~\eqref{eq:sga_formulation}, $t$ indexes the current frame. While SGG predicts the scene graph at frame $t$, SGA reads scene graphs up to frame $t$ to forecast the graph in future frame $t + n$.

\section{Our Proposed HyperGLM Approach}

\begin{figure}[!t]
\centering
\includegraphics[width=0.5\textwidth]{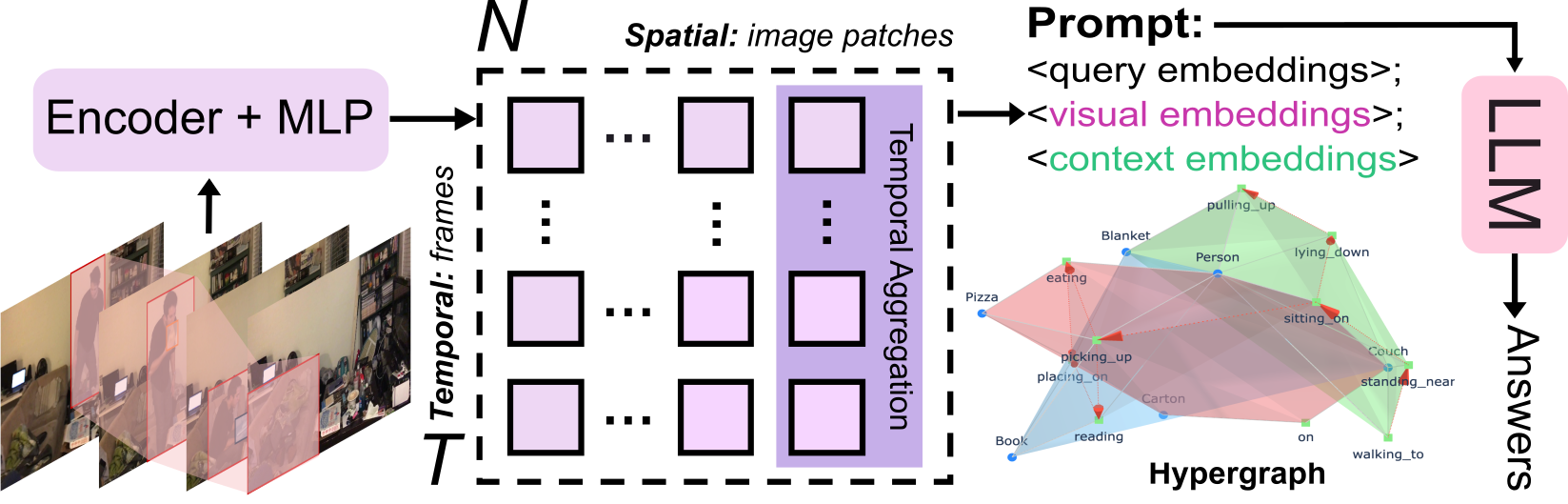}
\caption{Our HyperGLM framework comprises an image encoder, MLP projector, temporal aggregator, unified HyperGraph, and language model. It processes video frames by encoding each frame with the image encoder and MLP, extracting spatio-temporal features through image patch grids to generate $ N $ spatial tokens per frame. The temporal aggregator compresses the $ T \times N $ embeddings over time. The MLP projector then transforms these visual embeddings into the language feature space as frame tokens, interleaved with language tokens, and fed into the Large Language Models.} 
\label{fig:our_fw}
\vspace{-\baselineskip}
\end{figure}

In this section, we present our approach, which incorporates a unified HyperGraph into the LLMs, as illustrated in Fig.~\ref{fig:our_fw}. 

\subsection{Video Scene (Hyper)Graphs}\label{sec:vsh}
\begin{figure*}[!t]
\centering
\includegraphics[width=\linewidth]{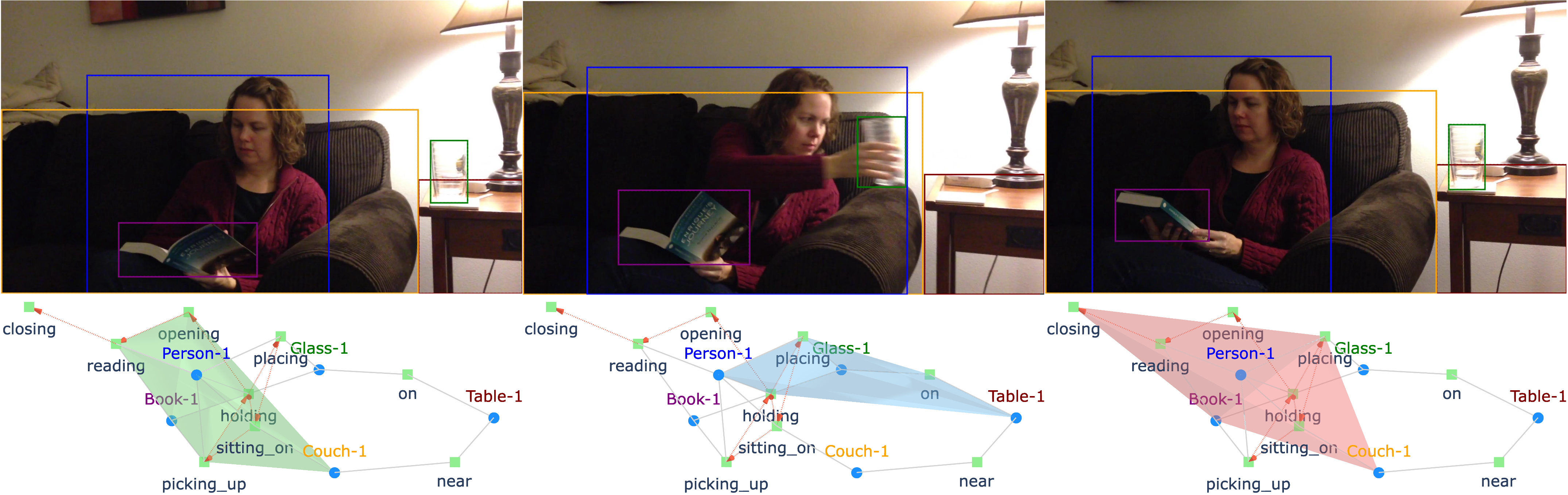}
\caption{Our Video Scene HyperGraph, including \textit{entity graphs} and a \textit{procedural graph}, as defined in~\cref{sec:vsh}. Blue nodes represent \textcolor{NavyBlue}{entities}, while green nodes denote \textcolor{ForestGreen}{relationships}. The entity graph captures spatial relationships (\textcolor{NavyBlue}{subject} \textcolor{Gray}{$\multimap$} \textcolor{ForestGreen}{relationship} \textcolor{Gray}{$\multimap$} \textcolor{NavyBlue}{object}), whereas the procedural graph models relationship transitions (\textcolor{red!75}{$\rightarrow$}). \textit{Hyperedges} are visualized as polygons, encapsulating interactions through chains of relationships. For instance, a \textit{hyperedge} illustrates a person picking up, holding, opening, and reading a book while sitting on a couch. HyperGraph is presented in 3D, see \textit{Supplementary video}.}
\label{fig:example_HyperGraph}
\vspace{-\baselineskip}
\end{figure*}

Traditional VidSGG methods~\cite{cong2021spatial, teng2021target, nguyen2024hig} construct scene graphs $\mathbf{G}_t$ to represent \textit{scene entities} and \textit{their relationships} as in Fig.~\ref{fig:compare}a. However, these graph-based approaches do not capture \textit{higher-order relationships with temporal dependencies} for video understanding~\cite{peddi2024towards}. To model this property, we propose a novel HyperGraph-based framework that constructs a unified HyperGraph $\mathcal{H} = (V_\mathcal{H}, E_\mathcal{H})$, representing spatial relationships within individual frames and temporal transitions across frames, illustrated in Fig.~\ref{fig:example_HyperGraph}.
HyperGraph extends traditional graph structures by allowing hyperedges to connect more than two nodes, making them particularly effective for modeling these higher-order relationships. To leverage this property, our unified HyperGraph (see Fig.~\ref{fig:compare}c) integrates \textit{entity scene graphs} $\mathbf{G}_t$ for each frame, capturing spatial interactions as introduced in~\cref{sec:prob_form}, along with a \textit{procedural graph} $\mathbf{P}$ that models their causal transitions, as detailed in~\cref{sec:vsh}. We unify these components using a \textit{random-walk HyperGraph construction}, which is defined in Alg.~\ref{alg:random_walk} to capture higher-order connectivity patterns and structural semantics, thereby approximating subgraph matching between the entity scene graphs and the procedural graph. This integration allows our approach to model current subjects' interactions and anticipate their future relationships.



\noindent \textbf{Procedural Graph Construction.}\label{sec:procedural-graph} SGA objective is to predict the set of relationships $E_{\mathbf{G}}$ in the next frame based on the current set of relationships. To model the temporal evolution of causal relationships between objects across video frames, we introduce a \textit{procedural graph} $\mathbf{P} = (V_{\mathbf{P}}, E_{\mathbf{P}})$, serving as the temporal counterpart to the entity scene graphs $\mathbf{G}_t$ as shown in Fig.~\ref{fig:compare}b. In particular, the procedural graph $\mathbf{P}$ can vary across datasets, modeling relationship transitions. 

Toward this goal, we model and denote the set of vertices in $\mathbf{P}$ as $V_{\mathbf{P}} = \{r_t^{ij} \in V_{\mathbf{G}} \mid 1 \leq t \leq T\}_{\neq}$, representing \textit{distinct relationship categories}. The set of edges in $\mathbf{P}$, \ie, $E_{\mathbf{P}} = \{ (r_m, r_n) \}$ represent possible \textit{causal transitions between these relationships}, where an edge $ (r_m, r_n) $ indicates that relationship $ r_m $ causally lead to relationship $ r_n $. We quantify causal transitions by calculating transition probabilities $w(r_m, r_n)$ via their observed frequencies as in Eqn.~\eqref{eq:transition_weights}.
\begin{equation}
    w(r_m, r_n) = \frac{\sum\limits_{t=1}^{T - 1} \sum\limits_{r^{ij}_{t, t+1}} \mathbbm{1}\left( r_t^{ij} = r_m \land r_{t+1}^{ij} = r_n \right)}{ \sum\limits_{t=1}^{T - 1} \sum\limits_{r^{ij}_{t, t+1}} \mathbbm{1}\left( r_t^{ij} = r_m \right)}
    \label{eq:transition_weights}
\end{equation}
where $\mathbbm{1}(\cdot)$ is the indicator function that counts transitions from relationship $r_m$ at current frame $t$ to $r_n$ at next frame $t + 1$. Next, self-loops $w(r_m, r_m)$ are removed from the graph, and these probabilities are normalized as in Eqn. \eqref{eq:transition_norm}.
\begin{equation}
    \sum_{r_n \in E_{\mathbf{P}}} w(r_m, r_n) = 1 \quad \text{for all } r_m \in E_{\mathbf{P}}
    \label{eq:transition_norm}
\end{equation}

By leveraging these probabilities, the procedural graph $\mathbf{P}$ enhances the prediction of future relationships. For each relationship $r_t^{ij}$ in frame $t$, the most probable relationship $r_{t+1}^{ij}$ in the next frame $t+1$ is determined as in Eqn. \eqref{eq:transition_probs}.
\begin{equation}
    r_{t+1}^{ij} = \arg\max_{r_n} P(r_n \mid r_t^{ij}, v^{i, j})
    \label{eq:transition_probs}
\end{equation}
where $P(r_n \mid r_t^{ij}, v^{i, j}) = w(r_t^{ij}, r_n) \times v^{i, j}$ is the probability of transitioning from relationship $r_t^{ij}$ to $r_n$, looking at object features. This allows the model to anticipate future interactions based on established temporal patterns.

\noindent\textbf{HyperGraph Construction.} To incorporate spatial relationships between objects in each frame and their causal temporal transitions, we construct a unified HyperGraph $\mathcal{H}$ that integrates the \textit{entity scene graphs} $\mathbf{G}_t$ and the \textit{procedural graph} $\mathbf{P}$. Specifically, a HyperGraph is an augmented representation combining the entity scene graphs with the procedural graph. This unified structure merges spatial and temporal relationships into a single graph, enabling the use of conventional graph algorithms while preserving the complex interactions captured by the HyperGraph. Mathematically, a unified HyperGraph $\mathcal{H} = (V_\mathcal{H}, E_\mathcal{H})$ is defined as in Eqn.~\eqref{eq:unified_HyperGraph}.
\begin{equation}
    \mathcal{H} = \left( \bigcup_{t=1}^{T} V_{\mathbf{G}_t} \cup V_{\mathbf{P}}, \quad \bigcup_{t=1}^{T} E_{\mathbf{G}_t} \cup E_{\mathbf{P}} \right)
    \label{eq:unified_HyperGraph}
\end{equation}
where $V_\mathcal{H}$ includes all entity nodes $v^{i}_t$ from each $\mathbf{G}_t$ and the relationship type nodes $V_{\mathbf{P}}$. The hyperedge set $E_\mathcal{H}$ includes pairwise relationships $E_{\mathbf{G}_t}$ within each $\mathbf{G}_t$, capturing spatial relationships and temporal transition edges $E_{\mathbf{P}}$ from $\mathbf{P}$, modeling the evolution of causal relationships across frames.


\noindent \textbf{Random-walk Algorithm.} We employ the random walks algorithm outlined in Alg.~\ref{alg:random_walk} to sample representative substructures from the unified HyperGraph $\mathcal{H}$, which captures connectivity patterns and mitigates the NP-hardness of exact subgraph matching. Specifically, these walks alternate between nodes and hyperedges, preserving the multi-node connections intrinsic to hyperedges and capturing the complexity of multi-object relationships and their transition across video frames. In each walk, a hyperedge $h_i$ aggregates the visited nodes, thereby encapsulating higher-order relationships. For example, in the entity scene graph $\mathbf{G}_t$, a walk might traverse from a ``person'' to a ``cup'' via the ``holding'', resulting in the hyperedge $h_i = \{\texttt{person}, \texttt{cup}\}$. Similarly, within the procedural graph $\mathbf{P}$, a walk might transition through a sequence of interactions such as ``holding'', ``placing'', and ``releasing'', forming the hyperedge $h_i = \{\texttt{holding}, \texttt{placing}, \texttt{releasing}\}$. Therefore, we generate sampled hyperedges $E_{\text{sampled}} = \{ h_i \mid i = 1, \dots, N_w \}$ as in L\ref{l:new_he} of Alg.~\ref{alg:random_walk} by conducting multiple random walks.

\begin{figure}[t]
\vspace{-\baselineskip}
\begin{algorithm}[H]
\caption{Random-walk for HyperGraph Construction.}\label{alg:random_walk}
\begin{algorithmic}[1]
\Require 
    $\mathcal{H} = (V_{\mathcal{H}}, E_{\mathcal{H}})$, Number of Walks $N_w$, Walk Length $N_l$
\Ensure 
    $\mathcal{H}' = (V_{\mathcal{H}}, E_{\mathcal{H}}')$
    
\State Initialize $E_{\text{sampled}} \gets \emptyset$

\For{$i = 1$ to $N_w$}
    \State Select $v_{\text{start}} \in V_{\mathcal{H}}$ uniformly at random
    \State Initialize walk sequence $S \gets [v_{\text{start}}]$
    
    \For{$j = 1$ to $N_l$}
        \If{ $j$ is odd} 
        \Comment{Node to HyperEdge}
            \State $v_{\text{current}} \gets S[j]$ 
            \State $\mathcal{E}_{v_{\text{current}}} \gets \{ h \in E_{\mathcal{H}} \mid v_{\text{current}} \in h \}$ 
            \If{$\mathcal{E}_{v_{\text{current}}} \neq \emptyset$}
                \State Select $h_j \in \mathcal{E}_{v_{\text{current}}}$ 
                \State Append $h_j$ to $S$
            \EndIf
        \Else
        \Comment{HyperEdge to Node}
            \State $h_{\text{current}} \gets S[j]$ 
            \State $V_{h_{\text{current}}} \gets h_{\text{current}}$ 
            \State Select $v_j \in V_{h_{\text{current}}}$
            \State Append $v_j$ to $S$
        \EndIf
    \EndFor
    
    \State $h_i \gets \{ v \in S \mid v \in V_{\mathcal{H}} \}$ \Comment{Form new HyperEdge}
    \If{ $h_i \notin E_{\mathcal{H}} \cup E_{\text{sampled}} $}
        \State $E_{\text{sampled}} \gets E_{\text{sampled}} \cup \{ h_i \}$
    \EndIf
\EndFor
    
\State $E_{\mathcal{H}}' \gets E_{\mathcal{H}} \cup E_{\text{sampled}}$ \label{l:new_he}
\State \Return $\mathcal{H}' = (V_{\mathcal{H}}, E_{\mathcal{H}}')$
\end{algorithmic}
\end{algorithm}
\vspace{-2\baselineskip}
\end{figure}

\subsection{Multimodal LLMs on HyperGraphs}

\noindent\textbf{Formulation.} Given an input video $\mathbf{V}$ and a task query $\mathbf{Q}$, we aim to generate a target answer sequence $\mathbf{A}$ of length $L$. Especially, by modeling a HyperGraph $\mathcal{H}$, the target answer $\mathbf{A}$ is generated via a process illustrated in Fig.~\ref{fig:our_fw}, defined as:
\begin{equation}
    p(\mathbf{A} | \keywordtri{\mathbf{V}}, \mathbf{Q}, \keywordtwo{\mathcal{H}}) = \prod_{i=0}^{L-1} p(x_i | \keywordtri{\mathbf{V}}, \mathbf{Q}, \keywordtwo{\mathcal{H}}, x_{<i})~\label{eq:MLLM_with_G}
\end{equation}
where $x_{<i}$ denotes the preceding token sequence, and $\mathbf{A} = [x_0, \dots, x_i, \dots, x_{L-1}]$ is the sequence of answer tokens reasoning video $\mathbf{V}$ by question $\mathbf{Q}$ and HyperGraph $\mathcal{H}$ in Fig.~\ref{fig:example_HyperGraph}.

\begin{figure*}[!t]
\centering
\includegraphics[width=\linewidth]{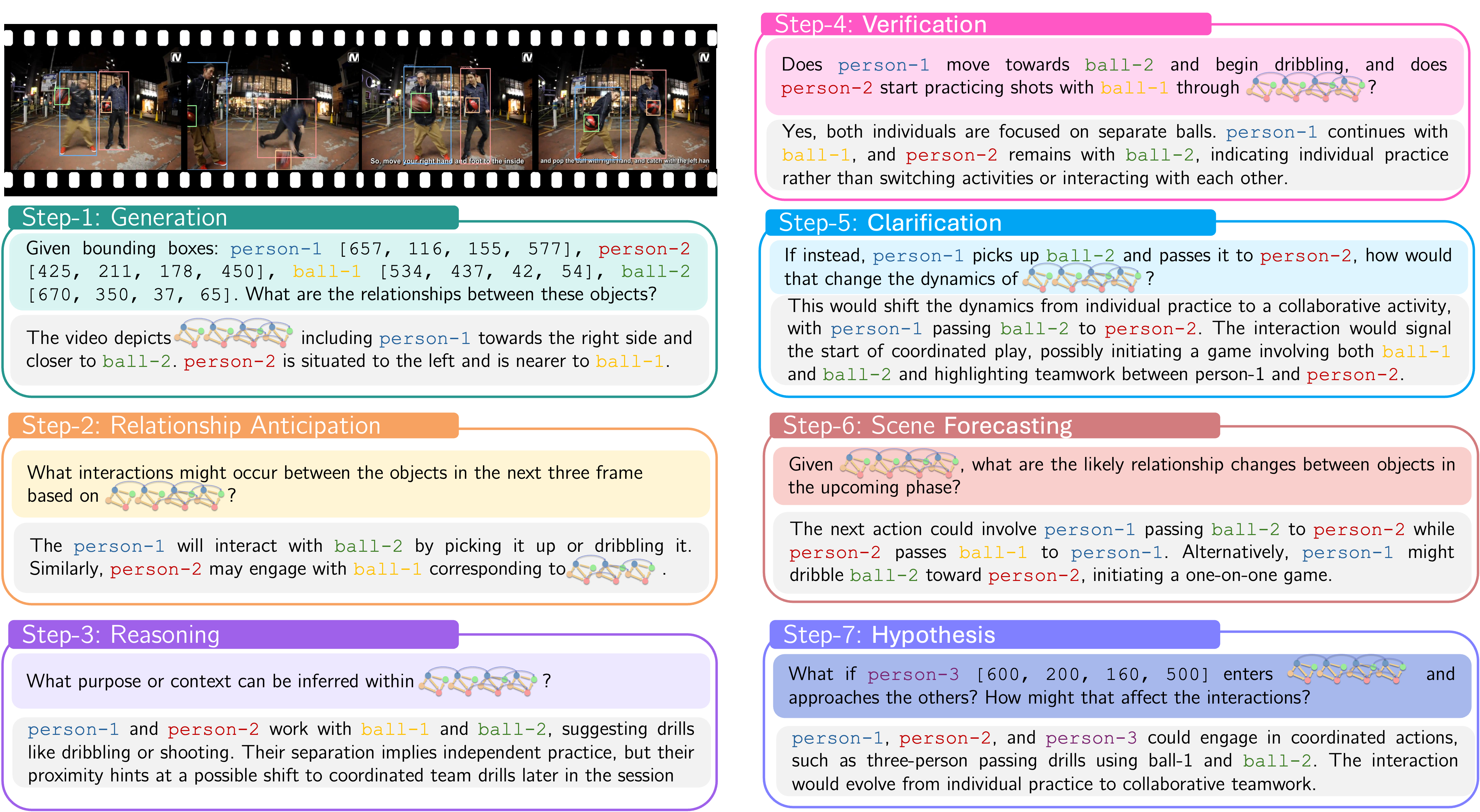}
\caption{An example of the diversified context within the streaming dialog in our VSGR dataset. \textbf{Best viewed in color and zooming in.}}
\label{fig:sample_conversation}
\vspace{-1.0\baselineskip}
\end{figure*}

As illustrated in Fig.~\ref{fig:sample_conversation}, our process begins with the \textit{generation}, constructing an entity scene graph $\mathbf{G}_t = (V^{e}_t, E_{\mathbf{G}})$ for each frame $t$ to capture detected objects and their relationships. \textit{Relationship anticipation} employs the procedural graph $\mathbf{P}$ to model temporal evolution and predict future interactions. The model then engages in \textit{reasoning} using the HyperGraph $\mathcal{H}$ and \textit{verification} through video captioning to ensure contextual relevance and accuracy. It refines understanding through \textit{ clarification}, generating answers by reasoning over the video and $\mathcal{H}$. In \textit{scene forecasting}, it predicts subsequent interactions by identifying new relationships and adapting to evolving contexts using transition probabilities in $\mathbf{P}$, anticipating future relationships $r_{t+n}^{ij}$ and improving conversational flow. Finally, the \textit{hypothesis} allows the model to propose and test conjectures about the underlying themes or intentions, synthesizing information from previous steps and leveraging $\mathcal{H}$ to explain patterns and forecast results.

\begin{table}[!t]
\centering
\caption{Comparisons of video scene graph datasets. SGG, SGA, VQA, VC, and RR represent Scene Graph Generation, Scene Graph Anticipation, Video Question Answering, Video Captioning, and Relation Reasoning.}
\label{tab:dataset_comparison}
\resizebox{\columnwidth}{!}{%
\begin{tabular}{@{}lccccccccc@{}}
\toprule
\multicolumn{1}{c}{\multirow{2}{*}{\textbf{Dataset}}} &
  \multicolumn{1}{l}{\multirow{2}{*}{\textbf{\#Frames}}} &
  \multicolumn{5}{c}{\textbf{Tasks}} &
  \multicolumn{3}{c}{\textbf{Annotations}} \\
\multicolumn{1}{c}{} &
  \multicolumn{1}{l}{} &
  \multicolumn{1}{c}{\textit{SGG}} &
  \multicolumn{1}{c}{\textit{SGA}} &
  \multicolumn{1}{c}{\textit{VQA}} &
  \multicolumn{1}{c}{\textit{VC}} &
  \multicolumn{1}{c}{\textit{RR}} &
  \multicolumn{1}{c}{\textit{Bbox}} &
  \multicolumn{1}{c}{\textit{Relation}} &
  \multicolumn{1}{c}{\textit{Text}} \\
\midrule
SportsHHI~\cite{wu2024sportshhi}     & 11.4K  &  \cmark &  \xmark &  \xmark &  \xmark &  \xmark &  \cmark &  \cmark & \xmark  \\
Action Genome~\cite{ji2020action} & 234.3K &  \cmark &  \cmark &  \xmark &  \xmark &  \xmark &  \cmark &  \cmark & \xmark  \\
AeroEye~\cite{nguyen2024cyclo}       & 261.5K &  \cmark &  \xmark &  \xmark &  \xmark &  \xmark &  \cmark &  \cmark & \xmark  \\
ASPIRe~\cite{nguyen2024hig}        & 1.6M   &  \cmark &  \xmark &  \xmark &  \xmark &  \xmark &  \cmark &  \cmark & \xmark  \\
PVSG~\cite{yang2023panoptic}          & 153K   &  \cmark &  \xmark &  \cmark &  \cmark &  \xmark &  \cmark &  \cmark & \cmark  \\
\rowcolor{tabhighlight} \textbf{VSGR (Ours)}          & \textbf{1.9M}   &  \cmark &  \cmark &  \cmark &  \cmark &  \cmark &  \cmark &  \cmark & \cmark  \\
\bottomrule
\end{tabular}%
}
\vspace{-\baselineskip}
\end{table}

\section{Video Scene Graph Reasoning Dataset}

\noindent\textbf{Limitations of Current Datasets.} Existing benchmarks (see Table~\ref{tab:dataset_comparison}) primarily support SGG and SGA, limiting their applicability to reasoning tasks. They suffer from inadequate support for VQA, VC, and RR, shallow annotations that fail to capture intricate object interactions, insufficient modeling of temporal dynamics, and poor multimodal integration. Thus, our VSGR dataset supports SGG, SGA, VQA, VC, and RR, enabling the reasoning capabilities of LLMs.

\subsection{Dataset Construction}

\noindent\textbf{Data Acquisition Stage.} We source videos from the ASPIRe~\cite{nguyen2024hig} and AeroEye~\cite{nguyen2024cyclo} datasets. While the ASPIRe dataset offers diverse, richly annotated videos emphasizing dynamic interactions and temporal changes, the AeroEye dataset offers drone-captured footage across various scenes.

\noindent\textbf{Comprehension Tasks via Question-Answering.} We introduce tasks that leverage fine-grained relationships from scene graphs, extending Scene Graph Generation to focus on relation understanding and subject/object interpretation using \texttt{<subject, relation, object>} triplets and leverage GPT-4/GPT-3.5 model for language generation.

\textit{Video Captioning (VC).} We generate 82,532 video-caption pairs, resulting in about 22 captions per video. Our process involves (1) extracting triplets from cropped video frames focusing on foreground objects, (2) generating background descriptions based on these triplets, and (3) combining the foreground triplets and background descriptions to produce captions. The average length is 893 characters, surpassing the PVSG~\cite{yang2023panoptic} dataset in quantity and detail.

\textit{Video Question Answering (VQA).} We develop 74,856 question-answer pairs by designing questions that explore diverse relationships, selecting subject and object categories to verify specific relationships, or choosing triplets to assess their uniqueness. This results in an average of approximately 20 questions per video, which exceeds existing benchmarks such as the MSRVTT-QA~\cite{xu2016msr} and MSVD-QA~\cite{xu2017video} datasets.

\textit{Relation Reasoning (RR).} Using the annotated scene graphs, we produce 61,120 relation reasoning tasks by selecting partial information as an incomplete input. Each task requires the model to deduce relationships among entities and identify the categories of the subject and object. With an average of approximately 16 tasks per video, our VSGR dataset provides a substantial collection for evaluating models' abilities in relational reasoning and scene understanding.

\noindent\textbf{Question and Answer Validation.} To ensure the quality and complexity of questions, we implement a rigorous validation process that combines generation by LLMs with human refinement. Initially, GPT-4/GPT-3.5 generates queries based on the \texttt{<subject, relation, object>} triplets extracted from the videos. Human annotators are trained with specific guidelines that emphasize clarity, relevance, and appropriate challenge levels, then review and refine questions. They enhance the questions by ensuring they are directly answerable from the video content and require careful reasoning about the depicted interactions and relationships.

We apply strict filtering criteria to improve the quality of the dataset. First, we eliminate questions that do not require video context and can be answered using general world knowledge, ensuring that models must rely on visual information from the videos. Second, we exclude questions that LLMs can answer correctly, increasing the challenge and utility of the dataset in evaluating advanced reasoning abilities. Finally, independent annotators perform a second round to review and evaluate the quality of the refined questions.

\subsection{Dataset Comparison}







As reported in Table \ref{tab:dataset_comparison}, our VSGR dataset represents a substantial advancement in video scene graph benchmarks. Our dataset comprises 3,748 videos and 1,841,243 frames, surpassing existing datasets in scale. Unlike other datasets that address only a limited subset of tasks, our dataset offers comprehensive task coverage, facilitating multifaceted evaluations of LLMs. In addition, our ground truth enriches relation annotations with comprehensive textual descriptions, enabling sophisticated reasoning and relationship predictions, as illustrated in Fig.~\ref{fig:sample_conversation}. Additionally, our VSGR dataset incorporates diverse viewpoints, including third-person, egocentric, and drone perspectives, enhancing its generalization. 

\section{Experiment Results}

\subsection{Implementation Details}

\noindent \textbf{Datasets.} We leverage our VSGR dataset across five tasks. In addition, we utilize the PVSG~\cite{yang2023panoptic} dataset for the SGG task and the Action Genome~\cite{ji2020action} dataset for the SGA task.

\noindent \textbf{Model Configuration.} We operate the CLIP-ViT-L-336~\cite{radford2021learning, dosovitskiy2021an} to encode each video frame into ten tokens (one CLS token and nine from 3$\times$3 average pooling). These tokens are fed into a two-layer MLP connector to the Mistral-7B-Instruct~\cite{jiang2023mistral} language model. For training, we apply LoRA~\cite{hu2022lora} to all linear layers with a rank of 128 and a scaling factor of 256, omitting vision-language alignment~\cite{liu2024visual}. We train for two epochs with a batch size of 128 over 16 iterations on 4 $\times$ GPUs, taking approximately six hours.

\noindent \textbf{Metrics.}\label{subsec:metrics}
We evaluate the SGG and SGA tasks using the Recall and mean Recall scores. In addition, we evaluate the VQA and RR tasks by Accuracy, Precision, Recall, and F1 scores. For the VC task, we utilize CIDEr, MENTOR, ROUGE-L, and BLEU-4 scores to validate our performance.

\noindent \textbf{Settings.}\label{subsec:settings} For the SGG and SGA tasks, we adopt the evaluation settings based on~\cite{ji2020action, peddi2024towards}. In particular, the model is provided with raw video frames and must detect objects using a pre-trained detector (\ie Faster R-CNN) and predict or anticipate their relationships. Especially for the SGA, we set the initial video input fraction ($\mathcal{F}$) to 0.9, following~\cite{peddi2024towards}.

\subsection{Ablation Study}\label{subsec:ablation}

\begin{figure}
\centering
\includegraphics[width=\linewidth]{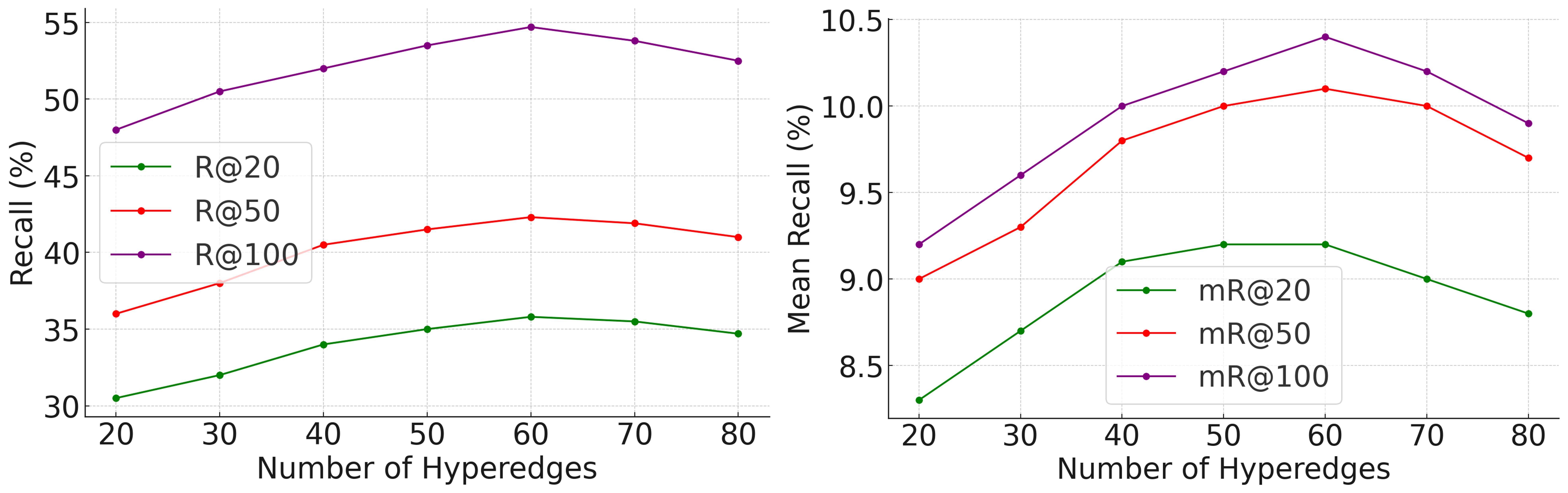}
\caption{Comparison of Recall (R) and mean Recall (mR) at different numbers of hyperedges for the SGG task on the VSGR dataset.}
\label{fig:nb_hyperedges}
\vspace{-0.5\baselineskip}
\end{figure}

\noindent\textbf{Hypergraph Parameters.} The number of walks $(N_w$) and walk length $(N_l$) are introduced in Alg.~\ref{alg:random_walk}, directly impacting the capacity to capture high-order relationships by determining the number of hyperedges. Although more hyperedges increase relational diversity, they can also introduce redundancy beyond an optimal point. Specifically, a higher $N_w$ broadens the range of sampled relationships, while a moderate $N_l$ balances depth. Our experiments demonstrate that optimal performances are achieved with 60 hyperedges ($N_w = 60$ and $N_l = 7$), shown in Fig.~\ref{fig:nb_hyperedges}. Further experiments on the SGG task are provided in Table~\ref{tab:ex_sgg}, and experiments on these parameters are included in the \textbf{Appendices}.

\begin{table}[!t]
\centering
\caption{Comparison (\%) on the VSGR and Action Genome datasets for the Scene Graph Anticipation (SGA) task at varying video input fractions $\mathcal{F}$.}
\label{tab:sga_ablation}
\resizebox{\columnwidth}{!}{%
\begin{tabular}{@{}c|l|ccc|ccc@{}}
\toprule
\multirow{2}{*}{$\mathcal{F}$} &
  \multicolumn{1}{c|}{\multirow{2}{*}{\textbf{Method}}} &
  \multicolumn{3}{c|}{\textbf{Action Genome}} &
  \multicolumn{3}{c}{\textbf{VSGR}} \\ \cmidrule(l){3-8} 
\multicolumn{1}{c|}{} &
  \multicolumn{1}{c|}{} &
  \textbf{R/mR@10} &
  \textbf{R/mR@20} &
  \textbf{R/mR@50} &
  \textbf{R/mR@10} &
  \textbf{R/mR@20} &
  \textbf{R/mR@50} \\ \midrule
\multirow{8}{*}{\textbf{0.3}} & STTran+~\cite{cong_et_al_sttran_2021}         & 13.9 / 3.5  & 21.6 / 7.3  & 40.8 / 20.3 & 12.0 / 4.0  & 19.0 / 8.0  & 35.7 / 18.0 \\
                              & DSGDetr+~\cite{Feng_2021}        & 14.3 / 3.6  & 21.8 / 7.6  & 41.3 / 21.2 & 12.5 / 4.2  & 19.5 / 8.3  & 36.0 / 19.0 \\
                              & STTran++~\cite{cong_et_al_sttran_2021}       & 15.4 / 6.2  & 27.2 / 14.1 & 48.6 / 32.2 & 14.0 / 6.5  & 22.0 / 11.0 & 39.0 / 25.1 \\
                              & DSGDetr++~\cite{Feng_2021}       & 16.8 / 8.4  & 29.0 / 16.7 & 48.9 / 32.3 & 14.5 / 7.0  & 22.5 / 12.0 & 40.0 / 26.0 \\
                              & SceneSayerODE~\cite{peddi2024towards}   & 23.3 / 13.3 & 32.5 / 20.1 & 45.1 / 33.0 & 18.0 / 9.5  & 26.0 / 15.0 & 42.0 / 30.0 \\
                              & SceneSayerSDE~\cite{peddi2024towards}   & 25.9 / 15.6 & 35.7 / 23.1 & 47.4 / 37.1 & 19.5 / 11.0 & 27.5 / 17.0 & 44.0 / 33.5 \\
                              & \textbf{HyperGraph (Ours)} & 26.5 / 14.8 & 36.2 / 22.1 & 49.3 / 37.2 & 18.8 / 9.3 & 27.2 / 16.3 & 42.5 / 27.4 \\
                              & \textbf{HyperGLM (Ours)} & \textbf{27.5} / \textbf{15.8} & \textbf{37.0} / \textbf{24.5} & \textbf{50.0} / \textbf{38.0} & \textbf{19.0} / \textbf{10.0} & \textbf{28.0} / \textbf{16.5} & \textbf{43.0} / \textbf{27.5} \\
\midrule
\multirow{8}{*}{\textbf{0.5}} & STTran+~\cite{cong_et_al_sttran_2021}         & 14.9 / 3.7  & 22.6 / 7.6  & 42.9 / 21.4 & 13.5 / 4.2  & 21.0 / 8.5  & 37.0 / 19.0 \\
                              & DSGDetr+~\cite{Feng_2021}        & 15.2 / 3.9  & 23.1 / 8.0  & 43.3 / 22.2 & 13.8 / 4.5  & 21.5 / 9.0  & 38.0 / 20.0 \\
                              & STTran++~\cite{cong_et_al_sttran_2021}       & 16.6 / 6.6  & 29.1 / 14.7 & 51.5 / 33.4 & 15.5 / 7.0  & 23.5 / 12.5 & 41.0 / 26.5 \\
                              & DSGDetr++~\cite{Feng_2021}       & 17.4 / 8.4  & 30.5 / 17.0 & 51.9 / 33.9 & 16.0 / 7.5  & 24.0 / 13.0 & 42.0 / 28.0 \\
                              & SceneSayerODE~\cite{peddi2024towards}   & 26.4 / 14.3 & 36.6 / 21.4 & 49.8 / 36.0 & 20.5 / 11.0 & 29.5 / 16.5 & 46.1 / 32.5 \\
                              & SceneSayerSDE~\cite{peddi2024towards}   & 28.4 / 16.3 & 38.6 / 25.1 & 51.4 / 39.9 & 21.5 / 12.5 & 31.0 / 18.5 & 48.0 / 35.7 \\
                              & \textbf{HyperGraph (Ours)} & 29.2 / 16.4 & 39.3 / 23.2 & 52.1 / 38.7 & 20.3 / 12.2 & 29.8 / 17.1 & 46.2 / 33.1 \\
                              & \textbf{HyperGLM (Ours)} & \textbf{30.0} / \textbf{17.0} & \textbf{40.5} / \textbf{27.5} & \textbf{53.5} / \textbf{40.5} & \textbf{21.5} / \textbf{11.5} & \textbf{31.5} / \textbf{18.5} & \textbf{46.5} / \textbf{30.0} \\
\midrule
\multirow{8}{*}{\textbf{0.7}} & STTran+~\cite{cong_et_al_sttran_2021}         & 16.6 / 4.2  & 25.1 / 8.5  & 47.2 / 24.0 & 15.0 / 5.0  & 24.0 / 10.5 & 41.0 / 21.5 \\
                              & DSGDetr+~\cite{Feng_2021}        & 16.8 / 4.3  & 25.3 / 8.8  & 47.4 / 24.7 & 15.5 / 5.2  & 24.5 / 11.0 & 42.0 / 22.0 \\
                              & STTran++~\cite{cong_et_al_sttran_2021}       & 19.0 / 7.7  & 32.8 / 17.1 & 56.8 / 36.8 & 17.0 / 8.0  & 27.0 / 14.0 & 45.0 / 29.0 \\
                              & DSGDetr++~\cite{Feng_2021}       & 19.8 / 9.5  & 34.1 / 19.2 & 56.7 / 37.2 & 17.5 / 8.5  & 28.0 / 15.0 & 46.1 / 30.0 \\
                              & SceneSayerODE~\cite{peddi2024towards}   & 32.1 / 16.5 & 42.8 / 24.4 & 55.6 / 39.6 & 23.5 / 13.0 & 33.5 / 19.0 & 51.0 / 36.0 \\
                              & SceneSayerSDE~\cite{peddi2024towards}   & 33.3 / 18.1 & 44.0 / 27.3 & 56.4 / 44.4 & 24.5 / 14.5 & 35.7 / 21.0 & 53.0 / 38.0 \\
                              & \textbf{HyperGraph (Ours)} & 34.3 / 19.2 & 45.2 / 25.3 & 57.2 / 42.1 & 22.3 / 13.2 & 34.2 / 19.3 & 50.4 / 35.3 \\
                              & \textbf{HyperGLM (Ours)} & \textbf{35.7} / \textbf{19.5} & \textbf{46.1} / \textbf{30.0} & \textbf{58.2} / \textbf{44.0} & \textbf{25.1} / \textbf{13.5} & \textbf{35.5} / \textbf{21.5} & \textbf{51.0} / \textbf{33.5} \\
\bottomrule
\end{tabular}%
}
\vspace{-1.0\baselineskip}
\end{table}

\begin{table}[!t]
\centering
\caption{Comparison (\%) on the VSGR and PVSG datasets for the Scene Graph Generation (SGG) task at Recall (R) and mean Recall (mR).}
\label{tab:ex_sgg}
\resizebox{\columnwidth}{!}{%
\begin{tabular}{@{}l|ccc|ccc@{}}
\toprule
\multirow{2}{*}{\textbf{Method}} & \multicolumn{3}{c|}{\textbf{PVSG}}                       & \multicolumn{3}{c}{\textbf{VSGR}}                       \\ \cmidrule(l){2-7} & \textbf{R/mR@20} & \textbf{R/mR@50} & \textbf{R/mR@100} & \textbf{R/mR@20} & \textbf{R/mR@50} & \textbf{R/mR@100} \\
\midrule
Transformer~\cite{yang2023panoptic}       & 4.0 / 1.8 & 4.4 / 1.9 & 4.9 / 2.0 & 25.7 / 6.3 & 34.5 / 6.5  & 43.5 / 7.0  \\
HIG~\cite{nguyen2024hig}               & 4.6 / 1.9 & 4.9 / 2.1 & 5.4 / 2.2 & 23.8 / 5.7  & 31.1 / 5.9  & 40.4 / 6.9  \\
CYCLO~\cite{nguyen2024cyclo}             & 5.8 / 2.0 & 6.1 / 2.2 & 6.7 / 2.3 & 29.4 / 7.1  & 36.4 / 7.7  & 47.7 / 7.7  \\
\textbf{HyperGraph (Ours)} & 6.5 / 2.2 & 7.0 / 2.4 & 7.5 / 2.6 & 31.6 / 7.8  & 38.8 / 8.3  & 50.3 / 8.5  \\
\textbf{HyperGLM (Ours)}   & \textbf{7.5} / \textbf{2.8} & \textbf{8.1} / \textbf{3.7} & \textbf{8.5} / \textbf{3.9} & \textbf{35.8} / \textbf{9.2}  & \textbf{42.3} / \textbf{10.1} & \textbf{54.7} / \textbf{10.4} \\ \bottomrule
\end{tabular}%
}
\vspace{-0.5\baselineskip}
\end{table}

\noindent\textbf{Video Input Fraction.} We adjust the initial video input fraction, $\mathcal{F}$, to 0.3, 0.5, and 0.7 for the SGA task. This adjustment allows the model to learn from varying observed portions and predict the unseen segment. Table~\ref{tab:sga_ablation} indicates that increasing the portion of the seen video improves performance, suggesting that additional visual context is beneficial. In addition, Table~\ref{tab:ex_sga} further confirms that the default setting at a higher input fraction ($\mathcal{F} = 0.9$) leads to optimal performance. We also present additional settings with varying video input fractions for the SGA task in the \textbf{Appendices}.

\subsection{Comparison with State-of-the-Arts}\label{subsec:sota}

\begin{figure*}[!t]
\vspace{-\baselineskip}
\centering
\includegraphics[width=\linewidth]{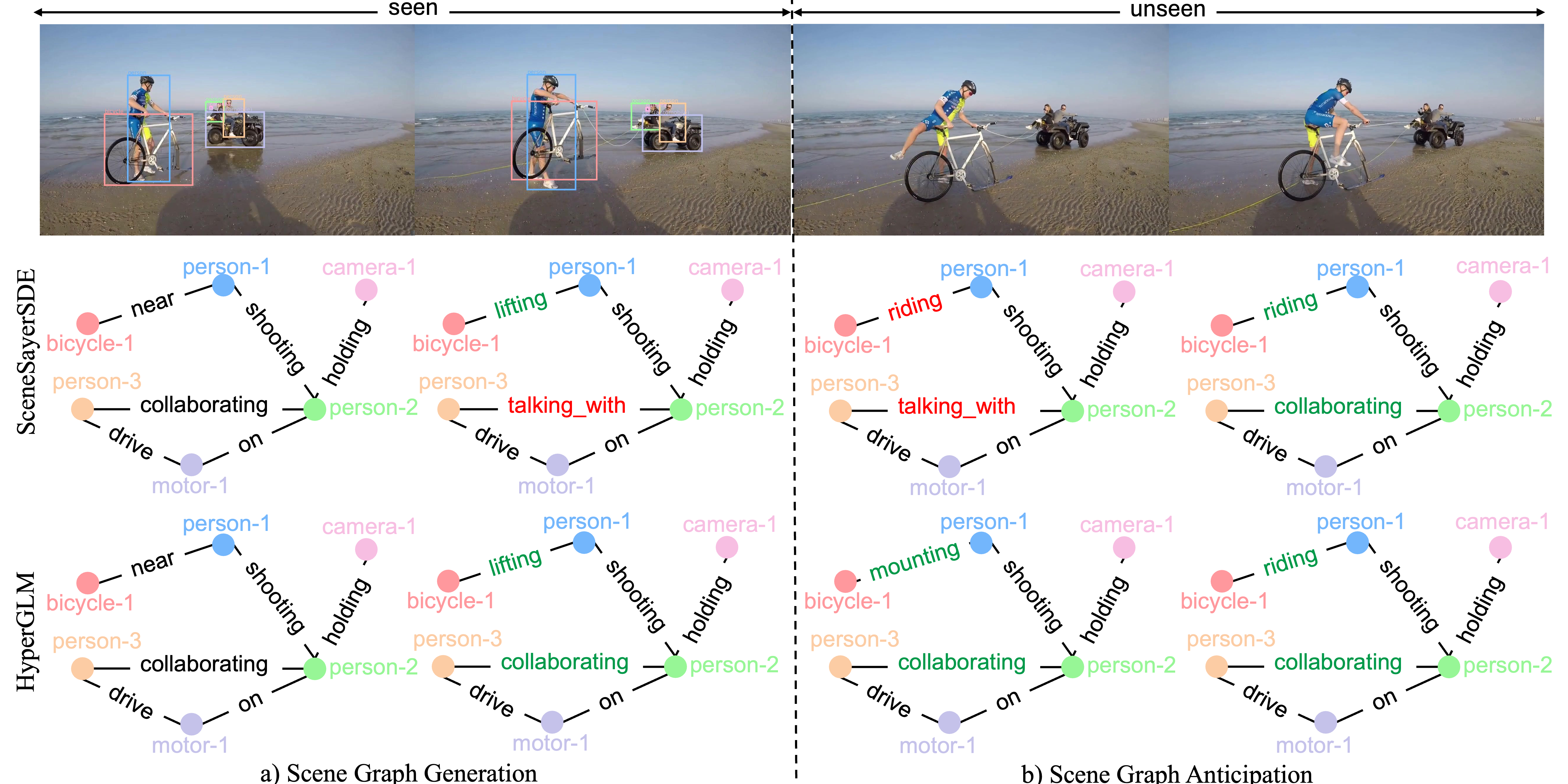}
\caption{Qualitative comparison of our HyperGLM approach versus SceneSayerSDE~\cite{peddi2024towards} for SGG and SGA. \textcolor{red}{Red} and \textcolor{ForestGreen}{green} edge labels denote incorrect and correct predictions, respectively. Our HyperGLM approach effectively captures the evolving interactions between \textcolor{lightblue}{\textbf{person-1}} and \textcolor{red!50}{\textbf{bicycle-1}} or \textcolor{green}{\textbf{person-2}} and \textcolor{wheat}{\textbf{person-3}} over time and anticipates interactions in unseen video frames, while SceneSayerSDE confuses to similar predicates. \textbf{Best viewed in color.}}
\vspace{-1.5\baselineskip}
\label{fig:prediction}
\end{figure*}

Table~\ref{tab:ex_sgg} demonstrates that HyperGLM significantly outperforms existing methods on the PVSG and VSGR datasets, achieving the highest R@20 scores of 7.5\% and 35.8\%, respectively. By leveraging hyperedges connecting multiple nodes within the HyperGraph, HyperGLM effectively captures complex object interactions and spatial dependencies, transcending traditional pairwise methods to generate more accurate and detailed scene graph representations. Notably, integrating a procedural graph within the HyperGraph \textit{reduces bias}. Our approach enhances mean Recall, reaching improvements of 2.8\% on the PVSG dataset and 9.2\% on the VSGR dataset, thereby addressing the long-tail distribution challenges that have struggled in previous methods~\cite{yang2023panoptic, nguyen2024hig, nguyen2024cyclo}. Furthermore, the LLM enhances the capacity of HyperGLM to infer and predict intricate relationships embedded within the unified HyperGraph, resulting in improved performance compared to HyperGraph, which shows a decrease of 4.2\% at R@20 on the VSGR dataset.
\begin{table}[!t]
\centering
\caption{Comparison (\%) on the VSGR and Action Genome datasets for the Scene Graph Anticipation (SGA) task at Recall (R) and mean Recall (mR).}
\label{tab:ex_sga}
\resizebox{\columnwidth}{!}{%
\begin{tabular}{@{}l|ccc|ccc@{}}
\toprule
\multicolumn{1}{c|}{\multirow{2}{*}{\textbf{Method}}} &
  \multicolumn{3}{c|}{\textbf{Action Genome}} &
  \multicolumn{3}{c}{\textbf{VSGR}} \\ \cmidrule(l){2-7} 
  \multicolumn{1}{c|}{} &
  \textbf{R/mR@10} &
  \textbf{R/mR@20} &
  \textbf{R/mR@50} &
  \textbf{R/mR@10} &
  \textbf{R/mR@20} &
  \textbf{R/mR@50} \\ \midrule
STTran+~\cite{cong_et_al_sttran_2021}         & 17.5 / 4.6  & 26.8 / 9.2  & 49.6 / 24.3 & 16.5 / 5.5  & 26.0 / 11.0 & 43.0 / 23.0 \\
DSGDetr+~\cite{Feng_2021}        & 17.9 / 4.7  & 27.7 / 9.7  & 51.4 / 25.9 & 17.0 / 6.0  & 27.0 / 11.5 & 44.5 / 24.5 \\
STTran++~\cite{cong_et_al_sttran_2021}       & 20.2 / 8.9  & 35.7 / 18.4 & 60.2 / 38.8 & 19.0 / 9.5  & 30.0 / 15.5 & 49.0 / 31.0 \\
DSGDetr++~\cite{Feng_2021}       & 22.2 / 11.4 & 37.1 / 21.0 & 61.0 / 39.5 & 19.5 / 10.0 & 31.0 / 16.0 & 50.0 / 32.5 \\
SceneSayerODE~\cite{peddi2024towards}   & 36.6 / 17.8 & 48.3 / 27.4 & 61.3 / 43.4 & 26.5 / 14.0 & 37.5 / 20.0 & 55.5 / 38.0 \\
SceneSayerSDE~\cite{peddi2024towards}   & 37.3 / 20.8 & 48.6 / 30.9 & 61.6 / 46.8 & 27.5 / 16.0 & 38.5 / 22.0 & 58.2 / 40.0 \\
\textbf{HyperGraph (Ours)} & 37.5 / 19.1 & 49.3 / 31.4 & 62.3 / 47.4 & 28.4 / 17.5 & 39.3 / 22.4 & 57.6 / 41.5 \\
\textbf{HyperGLM (Ours)} & \textbf{38.8} / \textbf{22.3} & \textbf{51.5} / \textbf{33.0} & \textbf{65.2} / \textbf{48.6} & \textbf{30.2} / \textbf{18.1} & \textbf{41.1} / \textbf{23.5} & \textbf{59.3} / \textbf{43.4} \\
\bottomrule
\end{tabular}%
}
\vspace{-1.5\baselineskip}
\end{table}

As shown in Table~\ref{tab:ex_sga}, our HyperGLM approach outperforms existing SGA methods on the Action Genome and VSGR datasets, achieving R@10 scores of 35.7\% and 25.1\%, respectively. This improvement stems from integrating procedural graphs that model \textit{causal relationships} within the HyperGraph structure. In contrast, the SceneSayer~\cite{peddi2024towards} method relies on NeuralODE and NeuralSDE to capture the latent dynamics of object interaction evolution. Our procedural graphs enable \textit{multi-step transitions} that explicitly capture the temporal evolution of relationships by modeling sequential interactions and their dependencies. In contrast, NeuralODE and NeuralSDE primarily focus on continuous-time dynamics, which may limit their effectiveness in handling discrete, multi-step relational changes as illustrated in Fig.~\ref{fig:prediction}b. Significantly, the HyperGraph is injected into the LLM, allowing the model to capture complex temporal patterns, resulting in better results than the HyperGraph.

\begin{table}[]
\centering
\caption{Comparison (\%) on VSGR for the  Video Question Answering.}
\label{tab:ex_vqa}
\resizebox{\columnwidth}{!}{%
\begin{tabular}{@{}l|cccc@{}}
\toprule
\multicolumn{1}{c|}{\textbf{Method}} &
  \multicolumn{1}{c}{\textbf{Accuracy}} &
  \multicolumn{1}{c}{\textbf{Precision}} &
  \multicolumn{1}{c}{\textbf{Recall}} &
  \multicolumn{1}{c}{\textbf{F1 Score}} \\ \midrule
Video-ChatGPT~\cite{maaz2023video}  & 33.2 & 35.1 & 32.3 & 33.6 \\
Video-LLaVA-7B~\cite{lin2023video} & 43.1 & 43.8 & 41.7 & 42.7 \\
MovieChat~\cite{song2024moviechat}      & 43.5 & 44.2 & 42.6 & 43.4 \\
Chat-UniVi-7B~\cite{jin2024chat} & 44.3 & 45.6 & 43.2 & 44.4 \\
\textbf{HyperGLM (Ours)}       & \textbf{45.4} & \textbf{47.2} & \textbf{44.3} & \textbf{45.7} \\ 
\bottomrule
\end{tabular}%
}
\vspace{-1.0\baselineskip}
\end{table}

\begin{table}[!t]
\centering
\caption{Comparison (\%) on VSGR for the  Video Captioning.}
\label{tab:ex_vc}
\resizebox{\columnwidth}{!}{%
\begin{tabular}{@{}l|cccc@{}}
\toprule
\multicolumn{1}{c|}{\textbf{Method}} & \textbf{CIDEr} & \textbf{MENTOR} & \textbf{ROUGE-L} & \textbf{BLEU-4} \\ \midrule
MV-GPT~\cite{seo2022end}        & 57.1 & 37.5 & 62.5 & 47.2 \\
CoCap~\cite{shen2023accurate}         & 54.3 & 29.4 & 61.8 & 42.8 \\
UniVL + MELTR~\cite{ko2023meltr} & 50.5 & 28.1 & 60.0   & 42.1 \\
\textbf{HyperGLM (Ours)}      & \textbf{54.5} & \textbf{30.7} & \textbf{64.9} & \textbf{48.8} \\ \bottomrule
\end{tabular}%
}
\vspace{-1.5\baselineskip}
\end{table}

In addition to the improvements in SGG and SGA shown in Fig.~\ref{fig:prediction}, Tables~\ref{tab:ex_vqa}, \ref{tab:ex_vc}, and \ref{tab:ex_rr} demonstrate the significant improvements using our HyperGLM approach in the VQA, VC, and RR tasks. In VQA, the hyperedges connect multiple objects, enabling HyperGLM to capture context-rich interactions more effectively, reaching an accuracy of 45.4\%. For VC, the HyperGraph structure supports evolving object connections over time, allowing our HyperGLM approach to generate captions that describe the relationship between objects and capture the temporal flow of interactions coherently, achieving scores of 54.5\% at CIDEr and 64.9\% at ROUGE-L. In RR task, the HyperGraph effectively manages intricate dependencies between multiple objects, resulting in precise relational inferences with an accuracy of 47.2\%.

\begin{table}[!t]
\centering
\caption{Comparison (\%) on VSGR for the  Relation Reasoning.}
\label{tab:ex_rr}
\resizebox{\columnwidth}{!}{%
\begin{tabular}{@{}l|cccc@{}}
\toprule
\multicolumn{1}{c|}{\textbf{Method}} & \textbf{Accuracy} & \textbf{Precision} & \textbf{Recall} & \textbf{F1 Score} \\ \midrule
Video-LLaVA-7B~\cite{lin2023video} & 41.3 & 42.5 & 40.2 & 41.3 \\
MA-LMM~\cite{he2024ma}         & 42.8 & 43.7 & 41.8 & 42.7 \\
LLaMA-VID-7B~\cite{li2025llama}   & 44.1 & 45.2 & 43.5 & 44.3 \\
\textbf{HyperGLM (Ours)}       & \textbf{47.2} & \textbf{48.4} & \textbf{46.5} & \textbf{47.4} \\ \bottomrule
\end{tabular}%
}
\vspace{-1.0\baselineskip}
\end{table}
\section{Conclusion}

In this paper, we have introduced HyperGLM, a novel VidSGG method that integrates scene hypergraph information into LLMs for context-aware and precise scene interpretation. Our approach effectively models complex interactions and higher-order relationships. It outperforms leading methods on benchmarks, including PVSG, Action Genome, and our newly collected VSGR dataset across five tasks.

\noindent \textbf{Limitations.} Although our HyperGLM approach effectively models multi-way interactions, managing many objects and their interactions can complicate relationship structures. As the HyperGraph expands, essential relationships may become obscured, reducing the clarity of scene interpretation. 


\noindent \textbf{Acknowledgment.} This material is based upon work supported by the National Science Foundation under Award No. OIA-1946391. We also acknowledge the Arkansas High-Performance Computing Center for providing GPUs.
 
{\small
\bibliographystyle{ieee_fullname}
\bibliography{content}
}



\renewcommand{\thesection}{\Alph{section}}
\renewcommand{\thefigure}{\Alph{section}.\arabic{figure}}
\renewcommand{\thetable}{\Alph{section}.\arabic{table}}
\renewcommand{\theequation}{\Alph{section}.\arabic{equation}}
\renewcommand{\thealgorithm}{\Alph{section}.\arabic{algorithm}}


\end{document}